# Learning Features from Co-occurrences: A Theoretical Analysis

**Yanpeng Li**
IBM T. J. Watson Research Center
Yorktown Heights, New York 10598
*liyanpeng.lyp@gmail.com*

## Abstract

Representing a word by its co-occurrences with other words in context is an effective way to capture the meaning of the word. However, the theory behind remains a challenge. In this work, taking the example of a word classification task, we give a theoretical analysis of the approaches that represent a word $X$ by a function $f(P(C|X))$, where $C$ is a context feature, $P(C|X)$ is the conditional probability estimated from a text corpus, and the function $f$ maps the co-occurrence measure to a prediction score. We investigate the impact of context feature $C$ and the function $f$. We also explain the reasons why using the co-occurrences with multiple context features may be better than just using a single one. In addition, some of the results shed light on the theory of feature learning and machine learning in general.

## 1      Introduction

In natural language processing (NLP) and information retrieval (IR), representing a word by its co-occurrences with contexts is an effective way to learn the meaning of the word and lead to significant improvement in many tasks [1][2][3]. The underlying intuition known as distributional hypothesis [4] can be summarized as: "a word is characterized by the company it keeps"[5]. However, there is a lack of theory to justify why it works, even for a simple task such as word classification or clustering. For example, why do the features from co-occurrences improve the performance of word classification or clustering? What types of context features or co-occurrence measures yield good performance? Why do the co-occurrences with multiple context terms perform better than a single one? The challenges lie in: 1) the general theory of feature learning is still an unsolved problem. 2) It is difficult for a theory to explain many different techniques that convert co-occurrences into prediction scores.

In this work, we first show that in a word classification task, simple co-occurrence with a single context feature can achieve perfect performance under some assumptions. Then we investigate the impact of context features and different co-occurrence measures without the assumptions. We also explain the reasons why using co-occurrences with multiple context features, e.g., vector representation can be better than just using a single one.

## 2      Notions

We consider a word classification task that assigns each word in a text corpus one or more class labels by the co-occurrences of the word with one or more context patterns. By going through the occurrence of each word in the corpus, we define each example as a tuple $(word, context, labels)$ denoted by $(X, \mathbf{c}, \mathbf{y})$. The component $X$ is a discrete random variable taking the values from the set $\{x_1, \dots, x_m\}$, where each element refers to a different word. The component $\mathbf{c} = (C_1, \dots, C_n)$ is a vector of features, where each component $C_k$ ($k$ is from 1 to

$n$) is a random variable that takes the value 1 or 0, indicating if a word or pattern appears in the context. Similarly we define $\mathbf{y} = (Y_1, ..., Y_o)$ as a vector of binary labels such as concepts. We aim to investigate how well we can assign each example a set of labels $\mathbf{y}$ by the co-occurrence information of $X$ and $C_k$ with insufficient or without $\mathbf{y}$ at training stage.

Note that for a label variable $Y_t$ ($t$ is from 1 to $o$), there are two cases. It may depend on both the current word $X$ and the context $\mathbf{c}$ or depend on $X$ only. In the tasks such as word sense disambiguation or part-of-speech tagging, the same word can be assigned to different labels in different contexts. In some tasks that investigate the meaning of words independently of contexts, e.g., building WordNet [6] or calculating word-word similarity [3], the same word $X$ can only be assigned to an unique label 0 or 1 for each $Y_t$. In other words, $Y_t$ is function of $X$, that is, for each $X$, $P(Y_t|X)$ equals 0 or 1. We consider both cases in our analysis, but we find that we can get much simpler results for the second case. Moreover, the word classification task for the second case can be generalized to the task of annotating everything with words. For example, describing an image with several words or a sentence is closely related to the task of classifying words into the class of description for the given image.

In the following sections, when we aim to investigate the impact of a single context feature $C_k$ or a single task $Y_t$, for simplicity, we use a random variable $C$ or $Y$ to denote $C_k$ or $Y_t$ respectively. So the tuple $(X, \mathbf{c}, \mathbf{y})$ can be simplified as $(X, C, Y)$ in the setting of single context feature and single task. In this work, we only consider the case of discrete random variables because text data as well as almost all natural data are discrete.

## 3   Conditional independence assumption

We show that in a special case if a context feature $C$ is conditionally independent with the word feature $X$ on label $Y$, the correlation coefficient between $P(Y=1|X)$ and $P(C=1|X)$ equals 1 or -1.

**Theorem 1.** Given random variables $X \in \{x_1, ..., x_m\}$, $C \in \{0,1\}$ and $Y \in \{0,1\}$, if

1. $P(C = 1, Y = 1) \neq P(C = 1)P(Y = 1)$

2. $P(C = 1, X = x_i|Y = 1) = P(C = 1|Y = 1)P(X = x_i|Y = 1)$ and $P(C = 1, X = x_i|Y = 0) = P(C = 1|Y = 0)P(X = x_i|Y = 0)$ for every $i$ from 1 to $m$

, we have:

$$Corr\big(P(Y = 1|X), P(C = 1|X)\big)^2 = 1$$

**Proof.** $P(C = 1|X) = \frac{1}{P(X)}(P(C = 1, X|Y = 1)P(Y = 1) + P(C = 1, X|Y = 0)P(Y = 0))$

$= \frac{1}{P(X)}(P(C = 1|Y = 1)P(X|Y = 1)P(Y = 1) + P(C = 1|Y = 0)P(X|Y = 0)P(Y = 0))$

(using the conditional independence assumption)

$= \frac{1}{P(X)}(P(C = 1, Y = 1)P(X, Y = 1)/P(Y = 1) + P(C = 1, Y = 0)P(X, Y = 0)/P(Y = 0))$

$= \frac{1}{P(X)}\left(\frac{P(C = 1, Y = 1)P(X, Y = 1)}{P(Y = 1)} + \frac{\big(P(C = 1) - P(C = 1, Y = 1)\big)\big(P(X) - P(X, Y = 1)\big)}{P(Y = 0)}\right)$

$= \left(\frac{P(C = 1, Y = 1)}{P(Y = 1)} - \frac{\big(P(C = 1) - P(C = 1, Y = 1)\big)}{P(Y = 0)}\right)P(Y = 1|X) + \frac{\big(P(C = 1) - P(C = 1, Y = 1)\big)}{P(Y = 0)}$

$= \frac{P(C = 1, Y = 1) - P(C = 1)P(Y = 1)}{P(Y = 1)P(Y = 0)}P(Y = 1|X) + P(C = 1|Y = 0)$

Since $P(C = 1, Y = 1) - P(C = 1)P(Y = 1) \neq 0$, $P(C = 1|X)$ is a linear function of $P(Y = 1|X)$, so we have $Corr\big(P(Y = 1|X), P(C = 1|X)\big)^2 = 1$

This theorem indicates that a perfect context feature for a given task is not necessarily the class label itself. Therefore, even if we have little information about $Y$, we still have the chance to know the information of $P(Y = 1|X)$ by $P(C = 1|X)$. Some previous works also reported the impact of conditional independence assumption in different tasks [7][8][9][10]. However, in practice it is almost impossible to get the case with exact conditional independence in different tasks. Even if it exists, it is difficult to find it because in order to calculate the conditional dependence we still need to know the joint probability $P(X, C, Y)$. Therefore, we need a theory to describe the performance in the situation with certain degree of conditional dependence. We also need to know the impact of different functions that convert simple co-occurrence measures e.g., $P(C = 1|X)$ to feature values or prediction scores. In the following sections, we will show the results for the cases.

## 4 Co-occurrences with a single context feature

In this section, we study the case of co-occurrence with a single context feature for a single task. Although it seems simple, there is still no systematic theoretical framework for it. Based on the theory about the function of discrete random variables, we analyze the impact of context features and the functions that convert co-occurrence measures to prediction scores. Note that the cases discussed in this section are not under the conditional independence assumption introduced in Section 3.

### 4.1 Function of a discrete random variable

Since the co-occurrence based learning actually converts the word variable $X \in \{x_1, \dots, x_m\}$ to another variable $S \in \{s_1, \dots, s_p\}$ by certain co-occurrence measure such as $P(X, C = 1)$ or $P(C = 1|X)$. In other words, there exists a function that maps $\{x_1, \dots, x_m\}$ to $\{s_1, \dots, s_p\}$, or equivalently, we can say $S$ is a function of $X$. Our goal is to find the function that achieves the best performance. Therefore, we need to investigate some general principles about the function of a discrete random variable at first. In this work, we use correlation coefficient of conditional probabilities as the measure of performance, because it tends to make the complex analysis simpler.

**Lemma 1.** Let $g$ be a function that maps $X \in \{x_1, \dots, x_m\}$ to another random variable $S \in \{s_1, \dots, s_p\}$, denoted by $S = g(X)$. For any binary random variable $Y \in \{0,1\}$ and any function $f$ that assigns $S$ a real number, we have the following results:

(1) $Cov(P(Y = 1|X), f(S)) = Cov(P(Y = 1|S), f(S))$
(2) $Cov(P(Y = 1|X), P(Y = 1|S)) = Var(P(Y = 1|S))$
(3) $Corr(P(Y = 1|X), P(Y = 1|S)) = \dfrac{\sqrt{Var(P(Y=1|S))}}{\sqrt{Var(P(Y=1|X))}}$
(4) $Corr(P(Y = 1|X), f(S)) = Corr(P(Y = 1|S), f(S)) Corr(P(Y = 1|X), P(Y = 1|S))$

**Proof.**
(1) $E(P(Y = 1|X)) = \sum_{i=1}^{m} P(X = x_i) \dfrac{P(X=x_i, Y=1)}{P(X=x_i)} = \sum_{i=1}^{m} P(X = x_i, Y = 1) = P(Y = 1)$
Similarly, we can prove that $E(P(Y = 1|S)) = P(Y = 1)$.
Therefore, we have $E(P(Y = 1|X)) = E(P(Y = 1|S))$
$Cov(P(Y = 1|X), f(S)) = Cov(P(Y = 1|X), f(g(X)))$ (by the definition of $g$)
$= \sum_{i=1}^{m} P(X = x_i) \Big( P(Y = 1|X = x_i) - E(P(Y = 1|X)) \Big) \Big( f(g(x_i)) - E(f(g(X))) \Big)$
$= \sum_{j=1}^{p} \Big( f(s_j) - E(f(S)) \Big) \sum_{g(x_i)=s_j} P(X = x_i) \Big( P(Y = 1|X = x_i) - E(P(Y = 1|X)) \Big)$
$= \sum_{j=1}^{p} \Big( f(s_j) - E(f(S)) \Big) P(S = s_j) \Big( P(Y = 1|S = s_j) - E(P(Y = 1|X)) \Big)$
$= \sum_{j=1}^{p} P(S = s_j) \Big( P(Y = 1|S = s_j) - E(P(Y = 1|S)) \Big) \Big( f(s_j) - E(f(S)) \Big)$
$\qquad\qquad\qquad\qquad\qquad \big( by\ E(P(Y = 1|X)) = E(P(Y = 1|S)) \big)$

$$= Cov(P(Y = 1|S), f(S))$$

(2) Since $P(Y = 1|S)$ is a function of $S$, let $f(S)$ be $P(Y = 1|S)$ and using (1) we have
$$Cov(P(Y = 1|X), P(Y = 1|S)) = Cov(P(Y = 1|S), P(Y = 1|S)) = Var(P(Y = 1|S))$$

(3) Based on (2), we have
$$Corr(P(Y = 1|X), P(Y = 1|S)) = \frac{Cov(P(Y = 1|X), P(Y = 1|S))}{\sqrt{Var(P(Y = 1|X))}\sqrt{Var(P(y|S))}}$$
$$= \frac{Var(P(Y = 1|S))}{\sqrt{Var(P(Y = 1|X))}\sqrt{Var(P(Y = 1|S))}} = \frac{\sqrt{Var(P(Y = 1|S))}}{\sqrt{Var(P(Y = 1|X))}}$$

(4)
$$Corr(P(Y = 1|X), f(S)) = \frac{Cov(P(Y = 1|X), f(S))}{\sqrt{Var(P(Y = 1|X))}\sqrt{Var(f(S))}}$$
$$= \frac{Cov(P(Y = 1|S), f(S))\sqrt{Var(P(Y = 1|S))}}{\sqrt{Var(P(Y = 1|S))}\sqrt{Var(f(S))}\sqrt{Var(P(Y = 1|X))}} \quad by\ (1)$$
$$= Corr(P(Y = 1|S), f(S))Corr(P(Y = 1|X), P(Y = 1|S)) \quad by\ (3)$$

The first equation plays a fundamental role, which indicates that if $S$ is function of $X$, to calculate the covariance between $P(Y = 1|X)$ and $f(S)$, if $S$ is available, we don't need the appearance of $X$ but only $S$ is enough, as if $X$ was forgotten or safely hidden. It also simplifies the analysis by avoiding explicit analysis of the divergence between $P(Y = 1|X)$ and $f(S)$ such as $|P(Y = 1|X) - f(S)|$ or $P(Y = 1|X)f(S)$. Therefore, the correlation coefficient can also be written in a much simpler form. In the word classification task, the term $f(S)$ can be viewed as a classifier based on the new features $S$. The $Corr(P(Y = 1|X), f(S))$ measures its correlation with the best possible classifier $P(Y = 1|X)$ based on the word features $X$ (single view) only. If $Y$ is a function of $X$ (the second case discussed in Section 2), we have $Corr(P(Y = 1|X), f(S)) = Corr(Y, f(S))$, which measures almost exactly the (regression) performance of the word classification task. Based on these results, we are able to analyze the performance of the new features generated by co-occurrences.

### 4.2   An upper bound of any function

**Theorem 2.** Given random variables $X \in \{x_1, \dots, x_m\}$, $C \in \{0,1\}$ and $Y \in \{0,1\}$, for any function $f$ that maps $P(C = 1|X)$ to a real number, there is
$$Corr\left(P(Y = 1|X), f(P(C = 1|X))\right)^2 =$$
$$Corr\left(P(Y = 1|P(C = 1|X)), f(P(C = 1|X))\right)^2 Corr\left(P(Y = 1|X), P(Y = 1|P(C = 1|X))\right)^2$$

**Poof**. Since $P(C = 1|X)$ is a function of $X$, let $S$ be the random variable that takes values $P(C = 1|X)$ and based on Lemma 1(4), we have:
$$Corr(P(Y = 1|X), f(S))^2 = Corr(P(Y = 1|S), f(S))^2 Corr(P(Y = 1|X), P(Y = 1|S))^2$$
Equivalently, we can write the equation as:

$$Corr\left(P(Y=1|X), f(P(C=1|X))\right)^2 =$$
$$Corr\left(P(Y=1|P(C=1|X)), f(P(C=1|X))\right)^2 Corr\left(P(Y=1|X), P(Y=1|P(C=1|X))\right)^2$$

It shows that the performance of the classifier $f(P(C=1|X))$ is determined by the product of two parts. The first part depends on both the context feature $C$ and the function $f$. The second part depends on the context feature $C$ only and not relevant to $f$. Therefore, if we fix the context feature $C$ to select the function $f$, using $Corr\left(P(Y=1|X), f(P(C=1|X))\right)^2$ (correlation with the gold standard) and $Corr\left(P(Y=1|P(C=1|X)), f(P(C=1|X))\right)^2$ (the "silver" standard) produce the same result. Similar to Lemma 1(1), we don't need the appearance of $P(Y=1|X)$ for function selection. In addition, since any correlation coefficient ranges from -1 to 1, we have:

$$Corr\left(P(Y=1|X), f(P(C=1|X))\right)^2 \leq Corr\left(P(Y=1|X), P(Y=1|P(C=1|X))\right)^2$$

It indicates that if we want to improve the performance by fixing a certain context feature $C$ and changing different functions $f$, e.g., from $P(C=1|X)$ to $\log\left(\frac{P(C=1,X)}{P(C=1)P(X)}\right)$ (the point wise mutual information), the performance cannot be arbitrarily high but upper bounded by the squared correlation coefficient between $P(Y=1|X)$ and $P(Y=1|P(C=1|X))$. Given fixed word feature set $X$ and label $Y$, the upper bound is determined only by the context feature $C$ but not relevant to the function $f$. Actually, the upper bound is a special case of maximum correlation coefficient [11][12]. If we want to build a high performing classifier $f(P(C=1|X))$, we must find a way to improve the upper bound, that is, to select a good context feature $C$. However, it is not easy to do it from this formula directly. Therefore, we relate the upper bound to the special case introduced in Section 3. Since $P(C=1|X)$ is a function of itself, based on Theorem 2, we have:

$$Corr(P(Y=1|X), P(C=1|X))^2 \leq Corr\left(P(Y=1|X), P(Y=1|P(C=1|X))\right)^2$$

The correlation coefficient $Corr(P(Y=1|X), P(C=1|X))^2$ not only reflects the performance of a special co-occurrence measure $P(C=1|X)$, but also provides a way to improve the upper bound (for other co-occurrence measures). Based on Theorem 1, under the conditional independence assumption, $Corr(P(Y=1|X), P(C=1|X))^2$ equals 1, so that $Corr\left(P(Y=1|X), P(Y=1|P(C=1|X))\right)^2$ equals 1 as well. As what we discussed in Section 3, it is important to know more about the cases out of the conditional independence assumption, but it is difficult to analyze $Corr(P(Y=1|X), P(C=1|X))^2$ based on the conditional dependence directly without any other assumption. In the following section, we show that we are able to obtain much simpler results by assuming that the label variable $Y \in \{0,1\}$ is a function of $X$.

### 4.3 Assuming that $Y$ is a function of $X$

In Section 2, we have pointed out that there are many cases in practice where $Y$ is a function of $X$. We believe that if we can explain everything about this task, we may naturally find the cues to move to more advanced tasks such as sentence classification and question answering.

**Theorem 3**. *If $Y \in \{0,1\}$ is a function of $X \in \{x_1, ..., x_m\}$, for a context feature $C \in \{0,1\}$, we have:*

$$Corr(P(Y=1|X), P(C=1|X))^2$$
$$= Corr(P(Y=1|Y), P(C=1|Y))^2 Corr(P(C=1|X), P(C=1|Y))^2$$

**Proof**. Since $Y$ and $C$ are both binary random variables, using Lemma 1 (4) replace $Y$ by $C$ without loss of generality, and we have:

$$Corr\big(P(C=1|X), f(S)\big)^2 = Corr\big(P(C=1|S), f(S)\big)^2 Corr\big(P(C=1|X), P(C=1|S)\big)^2$$

Since $Y$ is a function of $X$ and $P(Y=1|Y)$ is a function of $Y$, in the above equation replace $S$ by $Y$ and $f(S)$ by $P(Y=1|Y)$ respectively, and we have:

$$Corr\big(P(C=1|X), P(Y=1|Y)\big)^2$$
$$= Corr\big(P(C=1|Y), P(Y=1|Y)\big)^2 Corr\big(P(C=1|X, P(C=1|Y)\big)^2$$

Since $Y \in \{0,1\}$ is a function of $X$, there is $P(Y=1|X) = P(Y=1|Y) = Y$. Therefore, we have

$$Corr\big(P(Y=1|X), P(C=1|X)\big)^2 = Corr\big(P(Y=1|Y), P(C=1|X)\big)^2$$
$$= Corr\big(P(C=1|Y), P(Y=1|Y)\big)^2 Corr\big(P(C=1|X, P(C=1|Y)\big)^2$$

This result describes what a good context feature is and can give guidance for selection of context features for a particular task. It is much simpler and more practical than what we can get by the direct analysis of the conditional dependence, since we just need to know the joint distribution of $(X,C)$ and $(C,Y)$ rather than $(X,C,Y)$. This result can also be written as a simpler form as follows.

**Corollary 1.** *If $Y \in \{0,1\}$ is a function of $X \in \{x_1, \ldots, x_m\}$, for a context feature $C \in \{0,1\}$, we have:*

$$Corr\big(P(Y=1|X), P(C=1|X)\big)^2 = \frac{\big(P(C=1, Y=1) - P(C=1)P(Y=1)\big)^2}{P(Y=1)P(Y=0)Var\big(P(C=1|X)\big)}$$

**Proof**.
$$Corr\big(P(Y=1|X), P(C=1|X)\big)^2$$

$$= Corr\big(P(Y=1|Y), P(C=1|Y)\big)^2 Corr\big(P(C=1|X), P(C=1|Y)\big)^2 \qquad \text{(Theorem 3)}$$

$$= \frac{Cov\big(P(Y=1|Y), P(C=1|Y)\big)^2}{Var\big(P(Y=1|Y)\big)Var\big(P(C=1|Y)\big)} \frac{Var\big(P(C=1|Y)\big)}{Var\big(P(C=1|X)\big)} \qquad (Lemma\ 1(3))$$

$$= \frac{\big(E\big(P(Y=1|Y)P(C=1|Y)\big) - E\big(P(Y=1|Y)\big)E\big(P(C=1|Y)\big)\big)^2}{Var\big(P(Y=1|Y)\big)Var\big(P(C=1|X)\big)}$$

$$= \frac{\big(P(C=1, Y=1) - P(C=1)P(Y=1)\big)^2}{P(Y=1)P(Y=0)Var\big(P(C=1|X)\big)}$$

In the equation, the term $\big(P(C=1, Y=1) - P(C=1)P(Y=1)\big)^2$ describes the dependency between context feature $C$ and the class label $Y$. The term $Var\big(P(C=1|X)\big)$ addresses the dependency between context feature $C$ and each word feature $X$. We can see more clearly in another equivalent form:

$$Corr\big(P(Y=1|X), P(C=1|X)\big)^2 = \frac{P(Y=1)}{P(Y=0)} \frac{\left(\frac{P(C=1, Y=1)}{P(C=1)P(Y=1)} - 1\right)^2}{\sum_{i=1}^{m} \left(\frac{P(C=1, X=x_i)}{P(C=1)P(X=x_i)} - 1\right)^2}$$

From this mutual information style form, we can conclude that a good context feature for a particular task should be a trade-off between high dependence with label $Y$ and low dependence with every word feature $X$. The finding is similar to our previous work [10], but the result here is more general, under weaker assumptions, and more accurate (equation rather than inequality).

# 5 Co-occurrences with multiple context features

So far, we have investigated the co-occurrences with a single context feature only. In practice, we usually find representing a word by a vector of co-occurrences with multiple context features tends to perform better. In the following theorem, we show part of the reason by analyzing the performance of the vector from multiple context features.

**Theorem 4.** If we represent each $X$ by a $r$-dimensional vector $\mathbf{x} = \big(P(C_1 = 1|X), \ldots, P(C_r = 1|X)\big)$, where $C_1, \ldots, C_r$ are $r$ context features ($r \leq n$), for the label $Y \in \{0,1\}$ and any function $f$ that maps each vector $\mathbf{x}$ to a real number, we have

(1) $Corr\big(P(Y = 1|X), f(\mathbf{x})\big)^2 = Corr\big(P(Y = 1|\mathbf{x}), f(\mathbf{x})\big)^2 Corr\big(P(Y = 1|X), P(Y = 1|\mathbf{x})\big)^2$

(2) $Corr\big(P(Y = 1|X), P(Y = 1|\mathbf{x})\big)^2 = \dfrac{Corr\big(P(Y=1|X), P(Y=1|P(C_k=1|X))\big)^2}{Corr\big(P(Y=1|\mathbf{x}), P(Y=1|P(C_k=1|X))\big)^2}$ for every $k$ from 1 to $r$

**Proof.** (1) Since $X$ is a discrete random variable and $\mathbf{x} = \big(P(C_1 = 1|X), \ldots, P(C_r = 1|X)\big)$, we know that $\mathbf{x}$ is a discrete random variable and a function of $X$ as well. Therefore, based on Lemma 1 (4), we come to the conclusion directly:

$$Corr\big(P(Y = 1|X), f(\mathbf{x})\big)^2 = Corr\big(P(Y = 1|\mathbf{x}), f(\mathbf{x})\big)^2 Corr\big(P(Y = 1|X), P(Y = 1|\mathbf{x})\big)^2$$

(2) Since for each $\mathbf{x} = \big(P(C_1 = 1|X), \ldots, P(C_r = 1|X)\big)$ with a different value, the value of $P(C_k = 1|X)$ is unique, the discrete random variable $P(C_k = 1|X)$ is a function of $\mathbf{x}$. In Lemma 1 (3), replace $X$ by $\mathbf{x}$, and $S$ by $P(C_k = 1|X)$ without loss of generality, and we have

$$Corr\big(P(Y = 1|\mathbf{x}), P(Y = 1|P(C_k = 1|X))\big)^2 = \dfrac{Var\big(P(Y = 1|P(C_k = 1|X))\big)}{Var\big(P(Y = 1|\mathbf{x})\big)}$$

Since $\mathbf{x}$ is a function of $X$, and $P(C_k = 1|X)$ is a function of $X$, according to Lemma 1 (3), we have

$$Corr\big(P(Y = 1|X), P(Y = 1|\mathbf{x})\big) = \dfrac{Var\big(P(Y = 1|\mathbf{x})\big)}{Var\big(P(Y = 1|X)\big)}$$

$$Corr\big(P(Y = 1|X), P(Y = 1|P(c_k|X))\big)^2 = \dfrac{Var\big(P(Y = 1|P(C_k = 1|X))\big)}{Var\big(P(Y = 1|X)\big)}$$

By combining the above three equations, we have:

$$Corr\big(P(Y = 1|X), P(Y = 1|\mathbf{x})\big)^2 = \dfrac{Corr\big(P(Y = 1|X), P(Y = 1|P(C_k = 1|X))\big)^2}{Corr\big(P(Y = 1|\mathbf{x}), P(Y = 1|P(C_k|X))\big)^2}$$

Similar to Theorem 2, the performance of the vector-style representation can also be written as the products of two parts. The second part $Corr\big(P(Y = 1|X), P(Y = 1|\mathbf{x})\big)^2$ is an upper bound of the performance of any function, which is determined only by the set of context features $C_1, \ldots, C_r$. The result (2) shows that the upper bound from the vector is always better than the performance from any individual context feature, since we can prove that $Corr\big(P(Y = 1|X), P(Y = 1|\mathbf{x})\big)^2 \geq Corr\big(P(Y = 1|X), P(Y = 1|P(C_k = 1|X))\big)^2$. Similar to the proof of (2), it can be proved that $Corr\big(P(Y = 1|X), P(Y = 1|\mathbf{x})\big)^2$ always increases when more context features are introduced. However, $Corr\big(P(Y = 1|\mathbf{x}), f(\mathbf{x})\big)^2$ may decrease when the vector become longer, so its final performance $Corr\big(P(Y = 1|X), f(\mathbf{x})\big)^2$ may increase or decrease. Therefore, we need to find the factors that determine the combined performance of the two terms in the future.

# 6   Conclusions and future works

In this paper, we give a theoretical study of the approaches that learn the representation of words by the approaches like $f(P(C = 1|X))$ or $f(P(C_1 = 1|X), ..., P(C_r = 1|X))$. The theoretical framework is general enough to explain a set of approaches based on distributional semantics and is able to give guidance for algorithm design such as the selection of context feature and the co-occurrence metrics. In the next steps, we are going to give a deeper analysis of each formula that determines the performance, such as the diversity between multiple context features, find more principles that are constructive to algorithm design in practice and extend the theory to analyze other tasks in NLP and machine learning. Moreover, there are many important fundamental questions, e.g., what is the nature of co-occurrence? Is there some more fundamental reason for learning from co-occurrences? Is the creation of vector **x** related to the creation of the vector based instance representation in machine learning in general?